\documentclass[11pt]{article} 
\usepackage{hyperref} 
\usepackage[margin=1.5in]{geometry}

\usepackage{soul}

\usepackage{verbatim}
\usepackage{float}
\usepackage{graphicx}
\usepackage{epstopdf}
\usepackage{stmaryrd}
\usepackage{amsmath} 
\usepackage{placeins}
\usepackage{caption}
\usepackage{subcaption}

\pdfoutput=1 
\begin{document} 
\title{A neural circuit for navigation inspired by \\ \textit{C. elegans} Chemotaxis} 
\author{Shibani Santurkar and Bipin Rajendran \\ 
 \\\vspace{6pt} \hspace{-0.5in} Deparment of Electrical Engineering, 
Indian Institute of Technology Bombay, India \\
shibani@iitb.ac.in, bipin@ee.iitb.ac.in} 
\maketitle 
\begin{abstract} 
We develop an artificial neural circuit for contour tracking and navigation inspired by the chemotaxis of the nematode \textit{Caenorhabditis elegans}. In order to harness the computational advantages spiking neural networks promise over their non-spiking counterparts, we develop a network comprising $7-$spiking neurons with non-plastic synapses which we show is extremely robust in tracking a range of concentrations. Our \lq\lq{}worm\rq\rq{} uses information regarding local temporal gradients in sodium chloride concentration to decide the instantaneous path for foraging, exploration and tracking. A key neuron pair in the \textit{C. elegans} chemotaxis network is the ASEL \& ASER neuron pair, which capture the gradient of concentration sensed by the worm in their graded membrane potentials. The primary sensory neurons for our network are a pair of artificial spiking neurons that function as gradient detectors whose design is adapted from a computational model of the ASE neuron pair in \textit{C. elegans}. Simulations show that our worm is able to detect the set-point with approximately four times higher probability than the optimal memoryless L\'{e}vy foraging model. We also show that our spiking neural network is much more efficient and noise-resilient while navigating and tracking a contour, as compared to an equivalent non-spiking network. We demonstrate that our model is extremely robust to noise and with slight modifications can be used for other practical applications such as obstacle avoidance. Our network model could also be extended for use in three-dimensional contour tracking or obstacle avoidance.
\end{abstract} 

\begin{figure}
\centering
\includegraphics[width=3in]{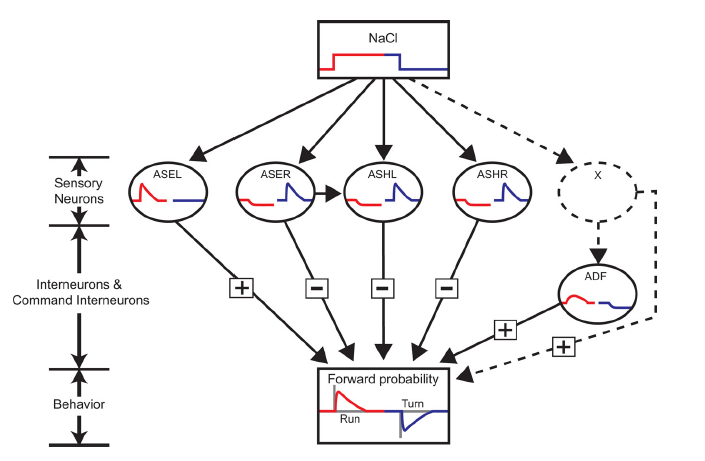}
\caption{ Neural network for chemotaxis towards high sodium chloride concentrations in \textit{C. elegans}. Red and blue curves show the responses of the neurons to NaCl up-steps and down-steps respectively. The black arrows signify speculated connections and the positive and negative signs show how the individual membrane potentials affect the final forward probability. As can be seen in the figure, there is high probability of turning when moving down a gradient and high probability of running forward while moving up a gradient. Reproduced with permission from \cite{Thiele23092009}}
\label{bionet}
\end{figure}

\section{Introduction} 
\ The human brain serves as an inspiration in the simplicity with which it is able to perform complex tasks consuming a fraction of time and power taken by even the most advanced computational systems today. However, the enormous complexity of the human brain which contains about 200 billion neurons interconnected with synapses in the order of trillions make the task of deciphering the algorithms employed for computation extremely challenging. Organisms with simpler connectivity statistics such as \textit{Caenorhabditis elegans} \cite{Brenner01051974}, Drosophila \cite{Konopka01091971}  etc. thus serve as ideal  platforms for experimental analysis. \textit{C. elegans} is considered a model organism in biology; its neural circuit comprising of merely $302$ neurons and about $5000$ chemical synapses, $2000$ neuromuscular junctions and $600$ gap junctions \cite{White12111986} is very well-understood today. In spite of the apparent simplicity,  the worm is capable of complex behaviors and abilities to learn, respond and adapt to various external chemical, thermal and mechanical stimuli. 

\ In this paper, we develop a spiking neural network (SNN) for contour tracking and navigation based on the concentration (or intensity) of the desired tracking variable, such as chemical concentration, which could find applicability in real word robotic applications. We have studied the NaCl chemotaxis circuit of  \textit{C. elegans} for inspiration to develop a bio-inspired circuit for navigation. Our aim is to develop circuitry that may eventually provide a more power-efficient and noise tolerant alternative to conventional algorithms, just as biological organisms prove to be much more efficient at computational tasks in terms of power and noise-tolerance than most advanced computing systems. In this context, the use of SNNs may be advantageous. For instance, it has been shown that SNNs have at least, the computational capabilities of previous generations of NNs, while requiring fewer number of neurons for performing many tasks  \cite{Maass19971659}. Information decoding in SNNs could be faster as they rely on the timing of individual spikes rather than average firing rate. The design and implementation of hardware technologies employing large SNNs are easier and potentially more power efficient since binary spikes control information transmission in an event-triggered manner. There is a thus significant impetus to develop SNN based circuits to tackle various engineering challenges in information processing, pattern recognition and navigation control. 

\ One of the key features of our navigation model is that it mimics the inherent gradient detection mechanisms used by the nematode. In the worm, this is achieved by the ASE pair of neurons whose output is an analog membrane potential which encodes information about the temporal gradient of the input chemical concentration. Gradient detection in our model is also performed using similar mechanisms. However, since our aim is to develop an intelligent network that uses only spikes for information transmission, we use a simple thresholding function at the output of our equivalent gradient detector neurons to create spikes that encode local gradient information. This is then used by downstream neurons in our model for making navigational decisions. We study the performance of our network for various concentration profiles as well as the implications of noise on our network\rq{}s performance. We also study the performance of our network in comparison with a non-spiking neural network for contour tracking. We show that the worm guided by the spiking network is much more efficient in navigating to and tracking a desired chemical concentration and also shows much better performance in noisy environments. Further we show that through a simple modification, our spiking network for contour tracking can be transformed to perform obstacle avoidance, another functionality that could have great applicability in real-world scenarios.

\ Previous works have focused on non-spiking neural circuits inspired by the chemotaxis network of \textit{C. elegans}, \cite{Shawn1999} propose a network model for chemotaxis of the \textit{C. elegans}, where it has been shown that linear networks comprising neurons with graded potentials are able to produce responses similar to the biological response of the nematode. They argue that their network strategies are designed to mimic \textit{klinotaxis} and \textit{klinokinesis}. \cite{Lockery2004} used simulated annealing  to identify networks that are capable of replicating the ideal sensorimotor transformations in the nematode. Through their simulations, a key three neuron network that acts as a differentiator circuit was identified, which is expected as the worm uses temporal gradient information for it\rq{}s navigation. \cite{PA} proposed a network model that relies on the conductance model of the ASE neurons to capture input to the network. Using a network with four neurons with graded potentials, three navigation strategies were analyzed - pirouettes, final turn angle control and steering and it was postulated that a combination of all the strategies yields best results with regards to mimicking biological chemotaxis characteristics.  

\cite{izquierdo2010evolution} use evolutionary algorithms to show that a simple network with merely one pair of OFF and ON neurons, which is chemosensory, and one pair of motor neurons is able to perform worm like \textit{klinotaxis} and is able to reproduce some key experimental observations, despite not being optimized to do so. They also provide a new theory for the underlying neural mechanism of \textit{klinotaxis} regarding the source of asymmetrical turning based on bilaterality of the sensory input. They claim that it could be a simple modulation of range of the sigmoidal inputs to the motor neurons, making one of the motor neurons sensitive to perturbations and the other insensitive. \cite{izquierdo2013connecting} combines neuroanatomical information about \textit{C. elegans} with simplifying assumptions about its network structure and environment, and with the help of genetic algorithms, a stochastic optimization is performed based on a measure of chemotactic performance. While they did not reward \textit{klinotaxis} in particular, the optimization produced an array of different networks that show worm-like \textit{klinotaxis} behaviour. They believe studying the characteristics of these networks will help design new experiments, the results of which can help highlight the relevant networks among the various possibilities. This information could serve as an additional constraint in the stochastic optimization producing a new set of possible networks. Such an iterative procedure could help in the advancement of both modelling and experiments. 

Significant work has been done by various research groups to find optimal navigational and tracking algorithms for many practical applications. \cite{kanayama1997new} develop an efficient algorithm for a robot under nonholonomic constraints to effectively track a fixed straight line, by developing an optimal steering function for determining the correction needed in the curvature of the path based on the error in the position and angle of the robot in addition to the current curvature of the robot\rq{}s path. They show that the robot\rq{}s path converges exponentially to the desired straight line after using linearization to find the optimal parameter set for the steering function and through the use of the \textit{Lyapupov} stability theory.  \cite{almeida1998comparing} compare various control strategies to effectively steer an autonomous robot to track a particular path. The authors compare both learning based strategies - such as self-organizing fuzzy logic based controllers (with and without integral control) and trained neural networks, to other simpler strategies such as controllers based on proportion (P), integral (I) or derivative (D) terms and fuzzy logic controllers. Among the simpler non-learning based approaches, they claim that a simple PD controller demonstrates the best performance, even surpassing a fuzzy logic based controller, but tuning a PD controller is extremely complicated and sometimes impossible. They claim that the self-organizing fuzzy network based approach is powerful due to it\rq{}s relatively low computational costs and quick learning abilities. Even though the neural network based approach considered produces the best results, it must be trained offline and often needs computational resources not available in frequently used micro-controller based systems. 

\cite{russell2003comparison} evaluates four reactive robot chemotaxis algorithms for applications that require tracking an air-borne gas or odor plume and locating its origin. In order to locate the plume, they compare passive monitoring, linear search and random walk based approaches. Subsequently, to track the plume they adopt control strategies inspired by \textit{E. coli} chemotaxis, as well as the silkworm moth, dung beetle and a gradient-based algorithm. They conclude that even though it is extremely simple, the \textit{E. coli} based strategy has limited applicability as it is easily perturbed by noise as compared to the other algorithms. While the silk worm based strategy performs well in rapidly fluctuating plumes, the gradient based strategy needs to be appended with a mechanism to prevent performance degradation in a turbulent environment. \cite{marques2002olfaction} also compare bacterial chemotaxis, silkworm moth algorithm and a gradient based approach for plume tracking, both using a simple gas sensor and an electronic nose. They show that while the gradient descent algorithm does perform better than the others, it does not do significantly better than the silkworm moth algorithm and that the results can be improved by using the electronic nose. \cite{kleeman1993thermal} compare different strategies to follow a particular thermal track with the help of a thermal sensor mounted on a robot. They compare gain based proportional and integral control strategies with a Kalman filter based approach and show that the later outperforms the former. 

The goal of our paper is to develop a noise-tolerant spiking neural circuit for use in robot navigation inspired by the key computational features of the \textit{C. elegans} network, which could be used to forage and identify specific chemical concentrations and track their contours, linear and non-linear, efficiently.

\section{Material \& Methods}

\subsection{Biological framework for chemotaxis of \textit{C.elegans}}
\label{subsec:bioevid}

\ One of the sophisticated abilities of the nematode \textit{C. elegans} is its ability to perform chemotaxis which is essentially movement prompted by chemical concentration, which is pivotal in its ability to find food, avoid danger, as well as other primal functions \cite{Ward01031973} \cite{Dusenbery01051973}. According to the data presented in \cite{Bargmann1991729}, about $32$ neurons in  \textit{C. elegans} are responsible for its chemotactic behavior to various water soluble and volatile compounds. In order to identify the neurons integral for chemotaxis, laser ablation experiments have been performed where specific neurons are killed and the behavior of the worm is observed after the ablation. According to \cite{Bargmann1991729}, these experiments have revealed that one specific neuron pair, the ASE neurons, when ablated cause the worm to have severely reduced chemotaxis towards water-soluble compounds like sodium chloride. However, if only the ASE pair is spared during an ablation, chemotaxis towards water-soluble attractants is preserved. This goes to show that the ASE neuron pair is crucial towards chemotaxis with residual functionality spread over numerous other neurons. Figure \ref{bionet} shows the neural network for NaCl in \textit{C.elegans} as proposed by \cite{Thiele23092009}.

In \textit{C. elegans}, most of the chemosensory neurons occur in symmetric pairs, i.e., in left-right neuron pairs. What makes the ASE neuron pair unique is that unlike other neuron pairs which tend to be functionally symmetric, i.e., both the neurons of the pair are excited by up-steps or down-steps of the chemical attractant, the ASE pair is functionally asymmetric. The ASEL neuron responds to up-steps in sodium chloride while the ASER responds to down-steps. 

The \textit{C. elegans} chemotaxis network guides the worm to a specific desired NaCl concentration, such as the cultivation concentration of the worm \cite{kun}. The worm employs two prime navigational strategies in order to convert the sensory cues into directed movement:  \\
 $\bullet$ {\textit{Klinokinesis}} or {\textit{biased random walk}}: The worm uses short-term memory about sodium chloride concentration to decide how it should navigate in the future. If the worm is moving  in a favorable direction, it makes rare turns or pirouettes. If on the other hand it has an unfavourable orientation, it makes frequent pirouettes \cite{Pierce-Shimomura01111999} \cite{Pierce-Shimomura15122005}. The result of this strategy is that it makes long runs in the correct directions and frequent turns away from the wrong directions. \\
$\bullet${\textit{Klinotaxis}}: The worm has a tendency to move towards the desired concentration  through sinusoidal motion which is skewed towards the desired attractant concentrations. \\
According to the data presented in \cite{kun}, when the worm is exposed to an environment with concentration equal to the desired set-point, it shows weakly negative \textit{klinotaxis} and exhibits no bias in \textit{klinokinesis}. 

In animal cells, action potentials are predominant where the frequency of the constant amplitude spikes capture information regarding the exciting signal. In contrast, most neurons in the worm generate graded action potentials which capture properties of the exciting signals in their shape and amplitude. 

\subsection{Modelling ASEL and ASER}
\label{sec:aselr}

\ One of the most interesting aspects of the actual chemotactic network in the worm is the ASE neuron pair \cite{Bargmann1991729} \cite{Thiele23092009} \cite{Miller30032005}. The ASE neurons  act as gradient detectors with the ASEL responding to up-steps and the ASER responding to down-steps in concentration. Most of the models presented in existing literature on bio-inspired neural networks based on \textit{C. elegans} chemotaxis capture information regarding the local sodium chloride concentration in a single current input given to a single sensory neuron and do not incorporate the ASE neuron pair. Our model for the ASE neuron pair is based on \cite{PA}. Not much is known  how exactly the local sodium chloride concentration is translated into potentials of the ASEL and ASER neurons. In the model presented in \cite{PA}, the external sodium chloride concentration is sensed by the worm through depolarizing and hyperpolarizing ion channels. The two ASE neurons are assumed to be independent of each other, i.e., there are no electrical or chemical synapses connecting the two neurons. The ASE neurons themselves are modeled by a simple conductance based approach.

The membrane potential of each of the ASE neurons  is modeled as
\begin{equation} \label{eq : pot}
\tau_m \frac{dV}{dt} = (V_0 - V) +  g^d (V_d - V) + g^h (V_h - V)
\end{equation}
where $\tau_m$ is the membrane time constant, $V_0$ is the resting membrane potential, $g^d$ and $g^h$ capture the conductivity of the depolarizing and hyperpolarizing ion channels respectively and $V_{d,h}$ represents the reversal potential of the respective ion channels. In general, the subscripts/superscripts $d/h$ represent depolarizing/hyperpolarizing channels.

The depolarizing ion channels are modeled by a three-state model comprising the unbound, bound and inactive states. The hyperpolarizing ion channels are modeled by a simpler two-state model with the unbound and bound states. The conductivity of both the channels is proportional to the fraction of channels in the bound state. The ion channels are by default in the unbound state and are non-conducting. The membrane potential of the ASE neurons is tied to the resting potential. 

For the depolarizing ion channels, transitions from the unbound to bound state are triggered when the local concentration exceeds (or goes below) some threshold concentration of the ASEL (ASER) neurons. These threshold concentrations are modeled subsequently as $NaCl_L$ and $NaCl_R$ for ASEL and ASER neurons respectively. The parameter $\alpha^d$, which determines the rate of transition from bound to unbound state also captures the adaptation property of the ASE neurons which is crucial to the worm\rq{}s ability to perform chemotaxis. The threshold concentrations adapt to the local sodium chloride concentrations, without which the worm would not be able to chemotax. Once in bound state, the depolarizing ion channels start conducting and hence the membrane potential of the ASE neurons starts increasing. Gradually, the depolarizing ion channels start transitioning from the bound state to the inactive state, which causes the membrane potential to fall due to a decrease in the conduction of the channels. Finally the ion channels return to the unbound state and the membrane potential returns to the resting potential. 

The fraction of the time the ion channel spends in the bound state will determine the peak value of the membrane potential, which is a graded potential as mentioned before. This is dependent on $\alpha^d$, i.e., the rate of transition from unbound to bound state, which is higher for greater changes in concentration. The equations governing the state transitions are  \begin{align}
\left[ \begin{matrix}
\dot{u^d} \\
\dot{b^d} \\
\dot{i^d}
\end{matrix}\right]
&= \left[ \begin{matrix}
- \alpha^d & \beta^d & \delta^d\\
\alpha^d & - \beta^d-\gamma^d & 0\\
0 &  \gamma^d & -\delta^d \end{matrix}\right]
\left[ \begin{matrix}
u^d \\
b^d  \\
i^d 
\end{matrix}\right]
 \end{align}
The hyperpolarizing ion channels are present only in the ASER neurons and serve to pull down membrane potential when concentration is greater that threshold concentration, $NaCl_R$. Normally, the hyperpolarizing ion channels are non-conducting. When the local concentration is greater than the threshold, these ion channels transition to the bound state and the conductivity of the hyperpolarizing channels increases thereby pulling down the membrane potential. Gradually the ion channels transition back to the unbound state and the potential returns to resting potential. Unlike the depolarizing ion channels, the hyperpolarizing ion channels do not adapt to the environmental sodium chloride concentration - $\alpha^h$ is  independent of local concentration (C) unlike $\alpha^d$. Their dynamics obey the equation
 \begin{align}
\left[ \begin{matrix}
\dot{u^h}  \\
\dot{b^h} 
\end{matrix}\right]
&= \left[ \begin{matrix}
- \alpha^h & \beta^h\\
\alpha^h & - \beta^h \end{matrix}\right]
\left[ \begin{matrix}
u^h \\
b^h  
\end{matrix}\right]
 \end{align}
Channel conductance is modeled as \begin{equation}
g^{d,h} = g_{max} \times ( b^{d,h} )^2 
\end{equation}
The transition rates $\alpha^h$, $\beta^{d,h}$ , $\gamma^{d}$ and $\delta^{d}$ have constant values. 

The magnitude of the response of ASEL/ASER to an up-step/down-step is determined by the rate $\alpha^d$, which resultantly depends on the change in concentration. It is modeled as follows. For the ASEL neuron, 
\begin{equation}
\alpha_L^d =
\begin{cases}
 \alpha_{L0}^d (C - NaCl_L) & if \  C \geq NaCl_L \\
 0 & \text{otherwise}
\end{cases}
\end{equation}
For the ASER neuron, 
\begin{equation}
\alpha_R^d =
\begin{cases}
 \alpha_{R0}^d (C - NaCl_R) & if \  C \leq NaCl_R \\
 0 & \text{otherwise}
\end{cases}
\end{equation}
$\alpha_{L0}^d $ and $\alpha_{R0}^d $ are scaling factors, $C$ represents local sodium chloride concentration sensed and $ NaCl_L$ and $ NaCl_R$ represent thresholds which adapt based on environmental  sodium chloride concentration. These equations ensure the graded quality of ASE neuron potentials since transition rate from unbound to bound state will be greater if the concentration deviates more from the threshold for the two neurons.

As mentioned previously, $\alpha^d$ captures adaptation of the worm to the ambient sodium chloride concentration. This adaptation is modeled as follows:
\begin{equation} \label{eq:asel_thresh}
\frac{dNaCl_{L}}{dt} = 
\begin{cases}
\frac{C - NaCl_{L}}{tau_{L}}  & if \  C \geq NaCl_L \\
\frac{-NaCl_{L}}{tau_L} & \text{otherwise}
\end{cases}
\end{equation}
\begin{equation} \label{eq:aser_thresh}
\frac{dNaCl_{R}}{dt} = 
\begin{cases}
\frac{C - NaCl_{R}}{tau_{R}}  & if \  C \leq NaCl_R \\
\frac{NaCl_{R}}{tau_R} & \text{otherwise}
\end{cases}
\end{equation}
We have modified the adaptation model in \cite{PA} for the ASER neuron.The threshold concentration, $NaCl_R$ has a possibility of getting stuck at 0 if it is exposed to an ambient with no NaCl. After that, the threshold will not be able to adapt to ambient concentration and ASER will be oblivious to the down-steps. Hence we must impose a minimum on the threshold value for the ASER neuron, i.e.,
\begin{equation}
NaCl_R = max(NaCl_R, NaCl_{R,min})
\end{equation}
where the value of $NaCl_{R}$ is dictated by  \ref{eq:aser_thresh}. 
The parameter $\alpha^h$ is modeled as
\begin{align}
\alpha_{L}^h &= 0 & \alpha_{R}^h &= \alpha_{0}^h H(C - \eta_{R})
\end{align}
where $\eta_R$ is the activation threshold for the hyperpolarising ion channels in the ASER neuron and $H(x)$ is the Heaviside function.  
\begin{figure}[t!]
        \centering
        \begin{subfigure}[b]{0.5\textwidth}
                \includegraphics[width=2.5in]{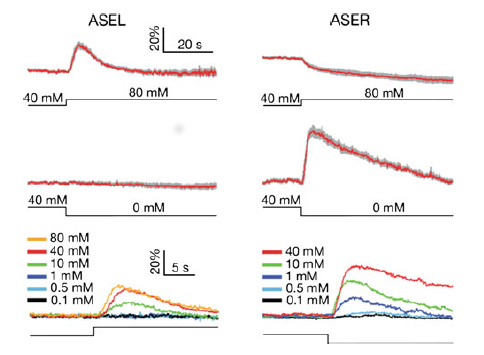}
                \caption{Calcium imaging data showing the membrane potential of ASE neurons when subjected to steps in NaCl concentration \cite{Suz}. }
                \label{biodata}
        \end{subfigure}%
\hfill
        ~ 
        \begin{subfigure}[b]{0.45\textwidth}
                \includegraphics[width=1.9in]{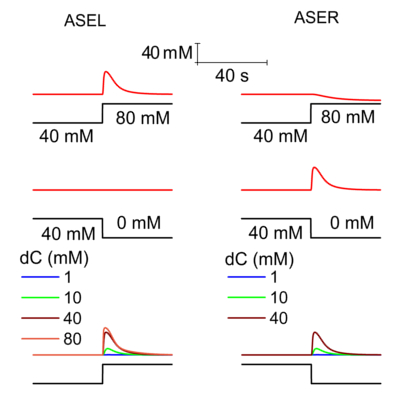}
                \caption{Membrane Potential of modeled ASE neurons to different steps in NaCl concentration. }
                \label{aselr_resp}
        \end{subfigure}
        \caption[Comparison of experimental response of ASE neurons in \textit{C.elegans} to numerical simulations]{\textit{Top}: Response of modeled ASE neuron pair to an up-step in NaCl concentration.
\textit{Middle}: Response of modeled ASE neuron pair to a down-step in NaCl concentration
\textit{Bottom}: Response of modeled ASE pair to steps of different magnitudes from a baseline concentration of $40\,$mM. The neurons respond more strongly for greater changes in concentration. The response of our numerically simulated neurons shows close resemblance to calcium imaging data for actual responses of ASE pair.}\label{fig:expnum}
\end{figure}

\subsection{Validating ASE neuron models}
Figure \ref{biodata} shows data from biological experiments to study the response of the ASE neurons to different concentration steps. The response of the modeled neurons when presented with similar concentration profiles is captured in Figure \ref{aselr_resp}. As can be seen, the response of the numerically simulated neurons show excellent agreement with the experimentally observed behavior when presented with steps in NaCl concentration, as in the figure.

Figure \ref{aselr_resp} shows the response of the simulated neurons when presented with different gradients in concentration. As is expected, the neurons respond much more strongly to sharper and stronger gradients than weak ones. Hence this model and our chosen parameters model the experimentally observed behavior of ASE neurons very well. 

\subsection{Modeling the Chemotaxis Network}
\label{sec:model}
 
 In standard chemotaxis,  \textit{C. elegans} navigates towards it\rq{}s cultivation concentration, or towards a concentration where it received food in the past. Likewise in real applications, it may be desirable for us to be able to determine the set point to which we would like the worm to navigate to and follow. So far we have modeled the ASE neurons to essentially be bio-mimetic. However as discussed before, these ASE neurons do not produce all-or-nothing action potentials but graded action potentials and hence a neural network comprising these neurons would rely on the analogue potential values for computation. 
 
\cite{Maass19971659}  have analyzed and compared the power of spiking neural networks with previous generations of neural networks to show that  the computational power of spiking neural networks is at least as high as first and second generation neural networks. They also show that in some cases spiking neural networks need fewer neurons to perform the same task computationally as previous generation neural networks. Further, constructing hardware implementations of large spiking networks is easier compared to non-spiking networks as all information transmission would be event-triggered and performed purely on the basis of binary spikes. Hence, we modify the ASE neurons to be more similar to animal neurons, i.e., have the spike rate encode information regarding input rather that the actual potential values, which may be more efficient computationally.  The spiking neuron pair inspired by the ASE neurons in \textit{C. elegans} could now be used together with other spiking neuron models, such as the Leaky Integrate and Fire neurons (LEIF) to develop a complete SNN to perform navigation and contour tracking. 

It has been ascertained by laser ablations that \textit{C. elegans} does not sense concentration gradients between its head and tail or between the left and right sets of neurons \cite{Rid}. The gradient in concentration that the worm senses can be perceived as a spatial gradient, between two positions of the worm or a temporal gradient. We use information regarding the temporal gradient in concentration to build our model.

The key features of the navigation control for our \lq\lq{}worm\rq\rq{} are discussed below. \\
$\bullet$ Ideally, when the  \lq\lq{}worm\rq\rq{} is on a roughly flat surface (in terms of concentration), away from the desired tracking concentration, it should explore the space by performing a random walk or foraging. This foraging should be done quickly, until  a \lq\lq{}favorable\rq\rq{} path/direction is identified.\\
$\bullet$ When the worm is moving up or down a gradient, in the direction away from the desired set-point, i.e., in an unfavorable direction,  the worm should alter its direction of motion. In our design, we chose to assign a clockwise turn  by $3.33^\circ$ when the worm is moving up the gradient and when it is already above the set-point. This translates to $dC/dt > 0$ and $ C > NaCl_{track}$, where $NaCl_{track}$ represents the set-point, $C$ represents the concentration sensed by the worm and $dC/dt$ represents the temporal gradient of this concentration. The worm makes a anti-clockwise turn in the opposite case with $dC/dt < 0$ and $ C < NaCl_{track}$. Switching the direction of turns in both the cases causes the worm to trace the contour in the opposite sense. We have selected clockwise and anticlockwise turns for the two cases (and not the same direction for turning) to ensure that the worm doesn\rq{}t keep circling the set-point.\\
$\bullet$ When the worm is moving in a favorable direction, i.e., $dC/dt > 0$ and $ C < NaCl_{track}$ or $dC/dt < 0$ and $ C > NaCl_{track}$ or $C = NaCl_{track}$, the direction of motion should be unaltered. The navigation model is inspired by the biological evidence for the navigation strategies used by \textit{C. elegans} as explained in Section \ref{subsec:bioevid} -- \textit{klinokinesis} or moving long distances without changing direction when it is on a \lq\lq{}favorable\rq\rq{} path and making frequent turns when it is not.\\

To build a spiking neural circuit that mimics these characteristic behaviors of the worm, we would need: \\
$\bullet$ A mechanism to ascertain if $C > NaCl_{track}$ or $C < NaCl_{track}$ \\
$\bullet$  A mechanism to determine the gradient - for which we can use the ASEL and ASER neurons \\
$\bullet$ A navigation mechanism which will tune the magnitude and direction of the velocity vector appropriately \\

\begin{figure}[t!]
        \centering
        \begin{subfigure}[b]{0.45\textwidth}
                \includegraphics[width=2.5in]{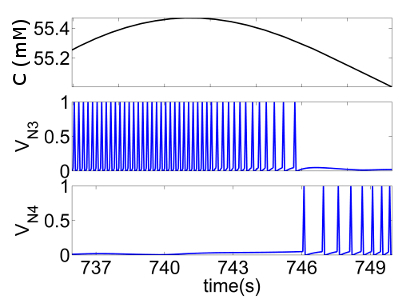}
                \caption{}
                \label{ase}
        \end{subfigure}%
\hfill
        ~ 
        \begin{subfigure}[b]{0.45\textwidth}
                \includegraphics[width=2.7in]{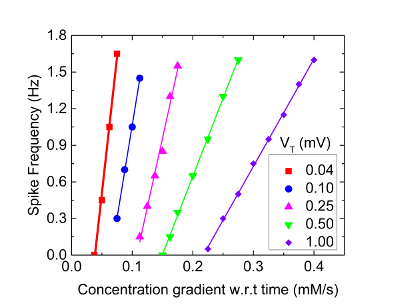}
                \caption{}
                \label{spike_freq}
        \end{subfigure}
        \caption[Response of designed artificial spiking neuron pair mimicing underlying neuron dynamics in ASE pair]{(a) Normalised response of gradient detector neurons, ASEL ($N_3$) and ASER($N_4$) when subjected to a certain NaCl profile.  ASEL spikes when the worm senses a positive gradient whereas ASER spikes when the worm senses a negative gradient. The spike frequency of the neurons is dependent on the gradient in concentration sensed as seen from the gradation in spike frequencies in the figure. (b) Spike frequency of the gradient detector neurons, i.e., ASE neuron pair  to different temporal gradients in NaCl concentration. By adjusting the threshold voltage $V_T$, it is possible to tune  the minimum temporal gradient that can be sensed by the gradient detectors. As threshold voltage decreases, neurons become more sensitive to lower gradients in concentration. The increase in spike frequency due to increase in concentration gradient becomes more pronounced at lower threshold voltages.}\label{fig:gradet}
\end{figure}

\subsection{Concentration Sensing Neurons}
\label{subsec:conc}

We need to have a neuron(s) that sense the ambient sodium chloride concentration and ascertain if $C > NaCl_{track}$ or $C < NaCl_{track}$. For this purpose we have two concentration sensing neurons, $N_1$ identifies if concentration is above the desired tracking concentration and $N_2$ identifies if concentration sensed is below it. These neurons are modeled as Leaky Integrate and Fire (LEIF) neurons, because of the simplicity of this model. The dynamics of the membrane potential $V(t)$ is governed by the equation
\begin{equation}
C \frac{dV(t)}{dt} = -g_L(V(t) -  V_0) + I_{app}(t) + I_{syn}(t)
\end{equation}
\begin{equation} 
\text{When} \  V(t)\geq V_T, V(t) \rightarrow V_{max}, V(t+dt) \rightarrow V_0.
\end{equation}
where the conductance and capacitance of the membrane is captured by C and $g_L$. $V_0$ and $V_T$ are the resting potential and the threshold voltage of the neuron respectively. Externally applied current is captured by $I_{app}(t)$  and synaptic current due to synaptic connections with other neurons is captured by $I_{syn}(t)$.

The synaptic current contribution due to a spike at time $t^k$ is given as
\begin{equation}
\label{eq:isyn}
I_s = I_0 \times w_{synapse}\times [e^{-(t-t^k)/\tau} - e^{-(t-t^k)/\tau_s}]
\end{equation} where $w_{synapse}$ denotes the strength of the synapse, and $\tau$ and $\tau_s$ are characteristic time constants. 

$N_1$ and $N_2$ are independent input neurons receiving only external input current and zero synaptic current.  The input current for $N_1$ is modeled as
\begin{equation}
I_{app}(t) = 
\begin{cases}
I_{app,0}  & if \  C > NaCl_{track} \\
 0 & \textit{otherwise}
\end{cases}
\end{equation}
The input current for $N_2$ is modelled as
\begin{equation}
I_{app}(t) = 
\begin{cases}
I_{app,0}  & if \  C < NaCl_{track} \\
 0 & \textit{otherwise}
\end{cases}
\end{equation}

Hence these two neurons have a fixed spike frequency and spike if concentration is greater or lesser than the tracking set-point respectively. As a result of the simplicity of these two neurons, our model is extremely versatile and it is extremely easy to tune the tracking set-point of the worm. 

\subsection{Gradient Detectors}

As mentioned previously, we would like to transform the model for ASEL and ASER neurons to develop spiking neuron models whose spike-frequency encodes information about the temporal gradient of the concentration. In order to do this, while computing membrane potential $V(t)$ for ASEL and ASER according to Equation \ref{eq : pot}, we apply the following additional constraint
\begin{equation}
\text{If} \ \ V(t) \geq V_T, V(t) = V_{max}, V(t+dt) = V_0
\end{equation}
In the overall circuit, ASEL and ASER are represented by $N_3$ and $N_4$ respectively. The response of the gradient detector neurons to a certain concentration profile is show in Figure \ref{ase}. As can be seen from the figure, the spike frequency depends on the concentration gradient sensed by the worm.

ASEL and ASER are very versatile gradient detectors and their minimum detectable temporal gradient  can be modulated by adjusting the threshold voltage $V_T$. This has great significance in the chemotaxis network as the minimum gradient that can be detected by the gradient detectors influences the worm behavior including how much the worm oscillates around the desired concentration. 

Figure \ref{spike_freq} captures the dependence of spike frequency of the modeled ASE neurons on the temporal derivative of concentration, for different $V_T$ values . We observe that as  the  threshold voltage decreases, the worm is able to sense smaller temporal gradients as expected. In addition, the spike frequency becomes more sensitive to the temporal gradient, and there is a greater change in spike frequency for the same change in temporal gradient. This provides a tuning mechanism to control the sensitivity and performance of the worm by simply controlling the threshold voltage of the ASE neurons.

\subsection{Detection of \lq\lq{}Unfavourable\rq\rq{} Orientation}

As discussed previously the worm needs to able to sense when it is moving in the \lq\lq{}wrong\rq\rq{} direction so that it can alter its direction. An unfavorable motion would be in the direction of positive gradient when the local concentration is already greater than $NaCl_{track}$ or moving in the direction of negative gradient when concentration is lower than $NaCl_{track}$. In order to detect these two cases, we use two LEIF neurons $N_5$ and $N_6$. 

Neuron $N_5$ is connected via excitatory synapses to neurons $N_1$ and $N_3$ and biased with a negative current $I_{bias,5}$, chosen to ensure that $N_5$ spikes if and only if both $N_1$ and $N_3$ spike. This implies that $N_5$ spikes only if local sodium chloride concentration is greater than $NaCl_{track}$ and the positive gradient detector is spiking. When $N_5$ spikes, the worm  makes a turn in the clockwise direction with a deterministic angle which we have chosen as $3.33^\circ$. Similarly, neuron $N_6$ is connected via excitatory synapses to neurons $N_2$ and $N_4$ and biased with an negative current $I_{bias,6}$, chosen to ensure that $N_6$ spikes if and only if both $N_2$ and $N_4$ spike. This implies that $N_6$ will detect the  other unfavorable case when the worm is moving down a gradient, away from the set point. When $N_6$ spikes, we stipulate that the worm makes an anti-clockwise turn with an angle of $3.33^\circ$.

The clockwise or anti-clockwise turns are determined by the spiking of $N_5$ and $N_6$, which in turn depend on the spiking of ASE neurons. An important feature of the ASE neurons is that the spike frequency is directly dependent on the temporal gradient of concentration. Therefore the worm would have a stronger tendency to turn if the deviation from the set-point is greater.

\subsection{Random Walk}

The worm should make rapid exploratory motion when it is \lq\lq{}lost\rq\rq{} and it is away from the set-point. This would happen when the worm is on a roughly flat concentration profile (or when the gradient is less than the detection threshold of the gradient detectors, i.e., ASE neurons) and is sensing a concentration which does not match $NaCl_{track}$. 

In our model, the worm makes decisions to random walk based on the spiking of $N_7$ which is implemented as an LEIF neuron. The worm can be considered \lq\lq{}lost\rq\rq{} when it is not at the desired set-point and it\rq{}s gradient detectors are not spiking, i.e., it is receiving no feedback whether it is on a favorable or unfavorable course. If the gradient detectors were spiking, the worm would know if was moving along a favorable or unfavorable orientation and would accordingly decide to keep moving straight or turn.

 Neuron $N_7$ is connected to $N_1$ and $N_2$ via excitatory synapses and $N_3$ and $N_4$ via inhibitory synapses. As a result, $N_7$ spikes in the presence of spikes of $N_1$ or $N_2$,  indicating the worm is away from the desired set-point and in the absence of spikes of $N_3$ and $N_4$, indicating that the worm is on a flat concentration profile, where the gradient sensed is less than detection threshold of $N_3$ and $N_4$ and hence it must random walk to find a favorable direction. This causes the worm to turn with a  randomly chosen angle from the interval $[-22.5^\circ, 22.5^\circ]$. 

\subsection{Velocity Model}
While foraging, i.e., when $N_7$ is spiking, the speed is chosen to be relatively high, $v_1 = 0.3\,$mm/s so that the worm can traverse a large area through rapid exploratory motions. In this phase it makes random turns with angles $[-22.5^\circ, 22.5^\circ]$. During tracking, i.e., when $N_5$ or $N_6$ spike, the worm makes a fixed turn with angle $3.33^\circ$ either clockwise or anticlockwise and moves with a lower velocity of $v_2 = 0.09\,$mm/s to improve tracking accuracy.  The velocity of the worm is thus either $v_1$ or $v_2$ depending on which neuron spiked last. If $N_7$ spikes, it sets velocity to $v_1$ whereas if $N_5$ or $N_6$ spike the velocity is set to $v_2$. In the absence of any spikes in $N_5$, $N_6$ and $N_7$ the velocity is determined by the last spiking neuron. 

\begin{figure}[!t]
\centering
\includegraphics[width=0.55\textwidth]{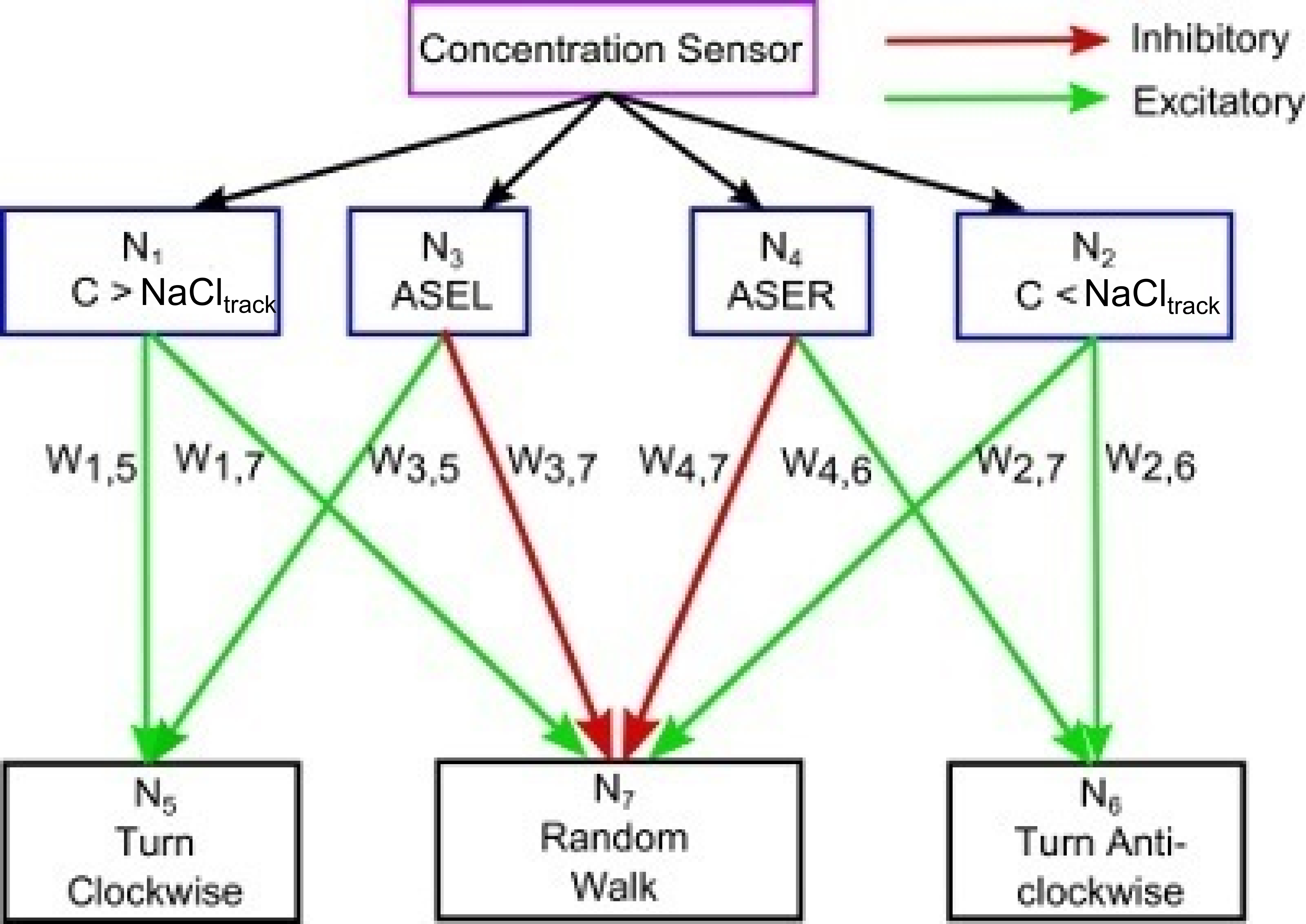}
\caption[Block diagram of spiking neural network for navigation and contour tracking inspired by \textit{C.elegans} chemotaxis]{Block Diagram of bio-inspired contour tracking network. Our network needs input from a single concentration sensor to steer the worm towards desired set-point $NaCl_{track}$. $N_1$ and $N_2$ are tuned to spike if the concentration is greater than or less than desired set-point. $N_3$ and $N_4$ are spiking neurons inspired by the ASE neuron pair in the C. elegans and behave as positive and negative gradient detectors respectively. $N_5$, $N_6$ and $N_7$ control navigation of the worm depending on spiking of $N_{1},N_2,N_3,N_4$.}
\label{circuit}
\end{figure}

 Figure \ref{circuit} captures the overall block diagram of our network.

\section{Results}
\subsection{Simulation Results}
\begin{figure}[t!]
        \centering
        \begin{subfigure}[b]{0.5\textwidth}
                \includegraphics[width=2.5in]{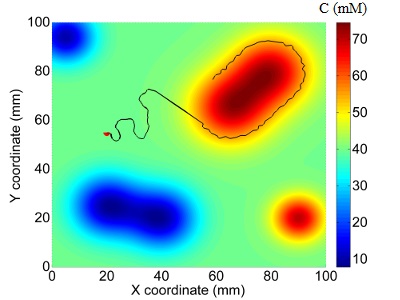}
                \caption{}
                \label{forage_55}
        \end{subfigure}%
\hfill
        ~ 
        \begin{subfigure}[b]{0.45\textwidth}
                \includegraphics[width=2.5in]{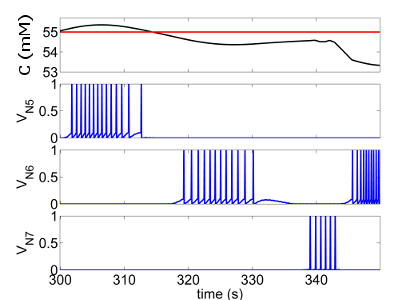}
                \caption{}
                \label{ons}
        \end{subfigure}
        \caption[Sample behaviour of model worm in identifying and tracking a specific target]{(a) Trajectory of our worm from initial position (indicated by red dot) at NaCl concentration of $40\,$mM to tracking desired set-point of $55\,$mM. The track shows initial random exploration by the worm. Once it finds a \lq\lq{}favourable\rq\rq{} direction, it keeps moving straight till the set-point is reached before starting tracking. In this case, the initial concentration sensed by the worm is lower than the desired set-point. (b) Spike patterns of $N_5$, $N_6$ and $N_7$ while tracking desired set-point. Figure shows $N_5$ spiking when worm moves up the gradient away from the set-point (deterministic clockwise turn), $N_6$ spiking when worm is moving down the gradient away from the set-point (deterministic anti-clockwise turn) and $N_7$ spiking when the worm senses a nearly flat profile and is lost, hence it must execute random walk. All membrane potential values are normalized.}\label{fig:gradet}
\end{figure}

In our simulations, the worm is placed on a $10\,$cm $\times 10\,$ cm  square plate with several hills and valleys of sodium chloride ranging in concentration from $10\,$mM to $70\,$mM. A typical navigation track for our worm is shown in  Figure \ref{forage_55}. The initial position of the worm is in a roughly flat region of the arena with concentration $40\,$mM, and the tracking set-point, $NaCl_{track}= 55\,$mM. As seen in the figure, the worm initially performs random exploratory motion to identify a favorable direction. The worm then travels straight, without pirouetting, till it reaches close to the desired set-point. Finally, the worm tracks the desired set-point with an accuracy of about $\approx 0.6\,$mM, which is $1\%$ of the range of concentration in the plate. Figure \ref{ons} shows the response of the output neurons, $N_5$, $N_6$ and $N_7$ to a certain concentration profile during this track. As can be seen from the figure, $N_5$ spikes when the worm is moving  away from set-point and $C > NaCl_{track}$. This will cause the worm to make a deterministic clockwise turn. Similarly, $N_6$ spikes when $C < NaCl_{track}$ and the worm is moving further away from set-point. This causes the worm to make a deterministic anti-clockwise turn. $N_7$ spikes when the worm is on an almost flat concentration profile and is away from the set-point causing the worm to random walk. 

\begin{figure}[t!]
        \centering
        \begin{subfigure}[b]{0.5\textwidth}
                \includegraphics[width=2.5in]{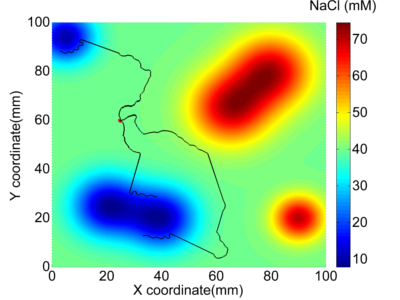}
                \caption{}
                \label{forage_20}
        \end{subfigure}%
\hfill
        ~ 
        \begin{subfigure}[b]{0.45\textwidth}
                \includegraphics[width=2.5in]{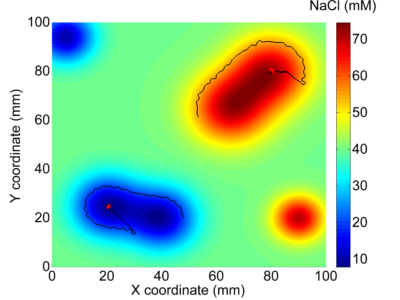}
                \caption{}
                \label{track_55}
        \end{subfigure}
        \caption[Model trajectories of worm in tracking various set-points]{(a)  Set of trajectories of the worm when it starts out from the initial set-point indicated by the red dot and tracks the desired concentration of $20\,$mM. In all three trajectories, the worm initially performs random exploration until it finds a favourable orientation. After that it moves straight until it reaches the desired concentration and then subsequently tracks the desired set-point (b) Simulated worm is able to locate and track set-point even if initial position (indicated by red dot) is at the peak of the hill (set-point being $55\,$mM) and at the bottom of the valley (set-point being $20\,$mM). Worm is able to detect and track the set-point irrespective of whether initial position has concentration greater or less than desired set-point.}\label{fig:gradet}
\end{figure}

Figure \ref{forage_20}  show three different tracks with the set-point at $20\,$mM which is in the valley. We simply change $NaCl_{track}$ to change our set point, every other parameter in the model remains the same. Figure \ref{track_55} shows two tracks, one with the worm tracking a set-point of $55$ mM with initial position at the top of the hill, and the other with the worm tracking set-point of $20$ mM starting out from the bottom of the valley. These figures show that the worm is successfully able to locate the set-point irrespective of whether the initial concentration sensed by the worm is higher or lower than desired set-point. 

\begin{figure}[!t]
\centering
\includegraphics[width=0.5\textwidth]{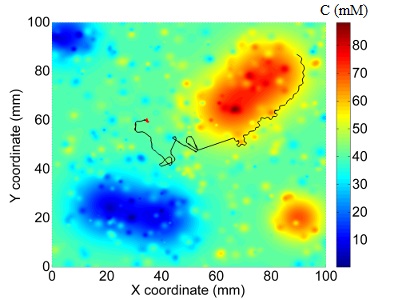}
\caption[Performance of network in presence of acute environmental noise]{Performance of the worm in a noisy environment demonstrating the noise-resilience of our algorithm. Track shows worm foraging and subsequently tracking a concentration of $55\,$mM in the presence of salt and pepper noise.}
\label{noise_ct}
\end{figure}

 \cite{Pappleby2013} show that using a network containing non-spiking ASE neurons based on \cite{PA}, the presence of noise severely disrupts the chemotactic response of their network.  In  Figure \ref{noise_ct}, the worm\rq{}s foraging and tracking behavior in an extremely noisy environment is portrayed. Despite the significant levels of noise, with absolute value in the range of $\approx 0-12\,$mM, the worm is able to track the contour effortlessly.  

\subsection{Scalability}

We now analyze how our model must change to operate in regimes with different temporal or spatial ranges. In real-world scenarios, such \textit{a priori} tuning of the network model may be needed depending on the environment it is deployed in. Our concentration sensing neurons are independent of these regimes since these neurons will spike at a fixed frequency if the local concentration is greater (for $N_1$) or lesser (for $N_2$) than the set point concentration. Hence to make these neurons sensitive to a different concentration gradient, in time or in space, we need not change anything.

 In order to increase the sensitivity of the gradient detectors, we must tune the threshold voltage $V_T$ of the ASE neuron pair as the value of $V_T$ determines the minimum temporal gradient of the concentration that  can be detected by the gradient detectors. As shown in Figure \ref{spike_freq}, the spike rate of the ASE neurons is a function of the $V_T$ and hence the weights of synapses $w_{35}$, $w_{46}$, $w_{37}$ and $w_{47}$ might need to be modified so that the functionality of $N_5$, $N_6$ and $N_7$ is preserved.

The implemented model can be easily tuned to operate in different spatial ranges. In our simulations the worm was exposed to hills and valleys in NaCl concentration over distances in the order of few centi-meters. To obtain similar performance at other length scales,  the velocity needs to be scaled in proportion so that the temporal gradient sensed by the worm is preserved. In order to operate over different temporal scales, we must merely modify the scale of all time-dependent parameters in the model, such as velocity and time constants appropriately.

\subsection{Performance Evaluation}

To evaluate the performance of our worm to identify a set-point by foraging, we perform several experiments where it starts from the same initial position with the tracking set-point set as $55\,$mM. For the $200\,$ simulations performed, the worm identified the set-point location in  $92\%$ cases within $1500\,$s. Further, we observed that in $60\%$ of the cases, our worm reaches the desired set-point in under $550\,$s. For the $92\%$ cases when it reaches the set-point within $1500\,$s, the average time needed to identify the set-point is $498.39\,$s with a standard deviation is $350.04\,$s. The deviation from the set-point, once the worm starts tracking the contour has mean value $0.6054\,$mM, which is $\approx 1\%$ of the total range of concentrations in the arena and a standard deviation of $0.078\,$mM, which is $0.13\%$ of the total range. In an extremely noisy environment, with the magnitude of noise ranging from $0-12\,$mM, our \lq\lq{}worm\rq\rq{} is able to track the set-point with the deviation from the set-point having average value $1.70\,$mM, $\approx 2.8\%$ of the range of concentrations in the arena and a standard deviation of $0.91\,$mM, or $1.5\%$ of the range. 

We also conducted a corner analysis to determine how our network would perform if  the sensitivity of our output neurons changed by varying the synaptic weights from the optimally chosen values by $10\%$, with all other network parameters kept unchanged. In these experiments,  the output neurons ($N_5$, $N_6$ and $N_7$) were either made more sensitive (spike more) or less sensitive (spike less), by inducing a $10\%$ drift in synaptic weights to study the impact on the foraging ability and tracking performance of our network. In order to make a neuron more sensitive, the weight of an incoming synapse would be increased, if it were an excitatory synapse and decreased (in magnitude) if it were an inhibitory synapse. Such tuning, where all the excitatory synapses of a neuron are strengthened (weakened) and all inhibitory synapses of a neuron are weakened (strengthened), would cause the neuron\rq{}s sensitivity to be increased (decreased) to the maximum extent possible while varying the synaptic weights. Although there are $2^6$ possible drift directions possible for the $6$ weights, the $8$ configurations we studied cover the worst case scenarios, where each of $N_5$, $N_6$ and $N_7$ was either made more or less sensitive to the maximum extent possible.

\begin{figure}[!t]
\centering
\includegraphics[width=4in]{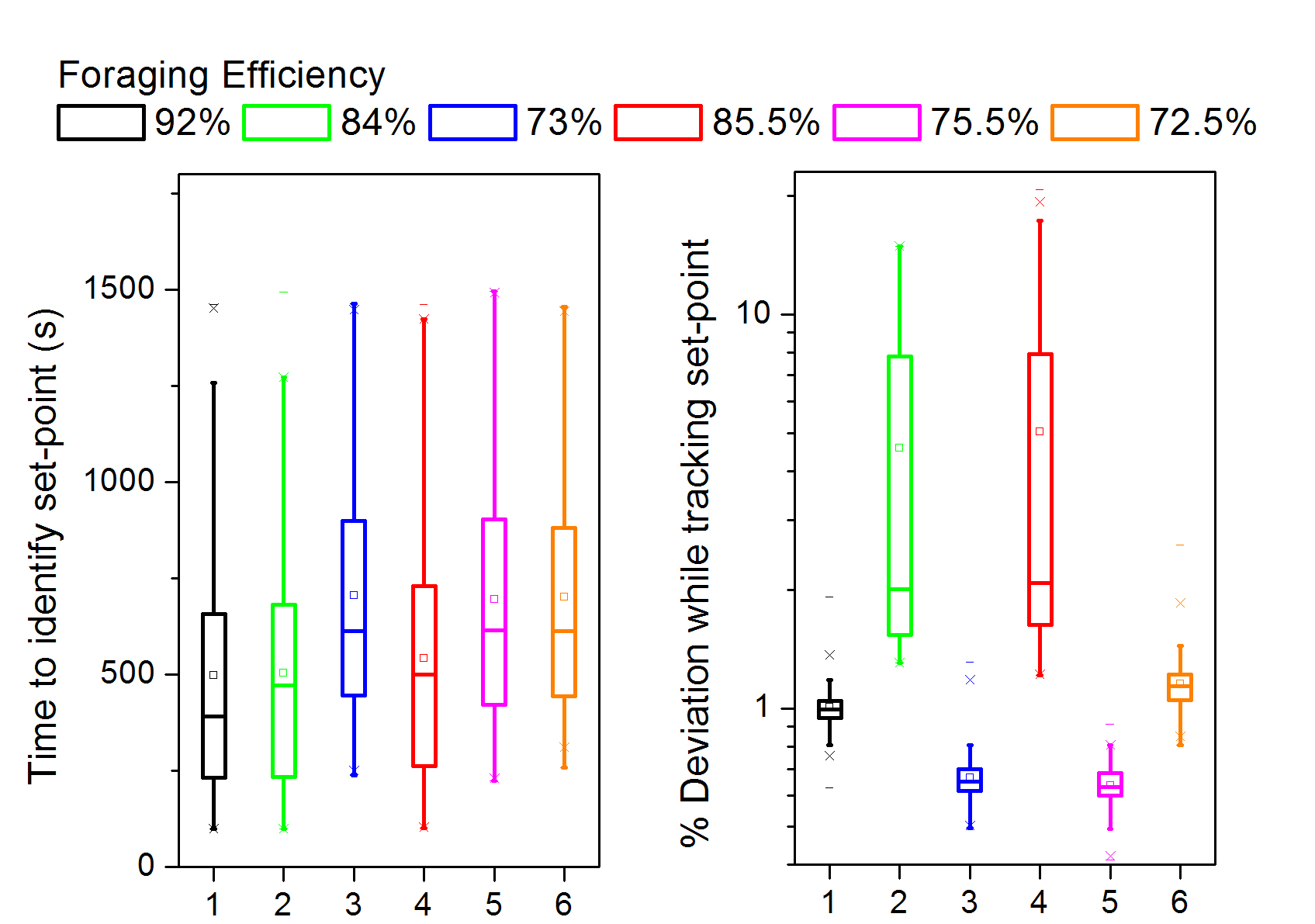}
\caption[Evaluation of resilience of network to parametric variations]{Performance evaluation of foraging and tracking behaviour of our network, (baseline in black) and corner cases when synaptic weights are altered by $10\%$ - (i) \textit{Green:} $N_5$, $N_6$ and $N_7$ are less sensitive (ii) \textit{Blue:} $N_5$, $N_6$ and $N_7$ are more sensitive (iii) \textit{Red:} $N_5$ \&$N_6$ are less sensitive and $N_7$ is more sensitive (iv) \textit{Pink:} $N_5$ \&$N_6$ are more sensitive and $N_7$ is less sensitive (v) \textit{Orange:} $N_5$ \&$N_7$ are less sensitive and $N_6$ is more sensitive. \textit{Left:} Time to identify set-point for the different network configurations \textit{Right:} \% Deviation (with respect to range of concentrations in the arena) while tracking set-point for different network configurations. Also shown in the figure is the foraging efficiency (percentage of 200 simulations where set-point is successfully identified) for the different network configurations.}
\label{pe}
\end{figure}

For each configuration of synaptic weights, we performed $200$ experiments each for $1500\,$s in which the worm had to identify and track a set-point of $55\,$mM. Figure \ref{pe} shows the performance, in foraging and tracking the set-point, of $5$ corner cases out of the total $8$ simulated cases which showed maximum deviation as compared to the baseline (\textit{Case 1}) which is our default network configuration. The five corner cases reported are - (i) \textit{Case 2:} $N_5$, $N_6$ and $N_7$ are less sensitive (ii) \textit{Case 3:} $N_5$, $N_6$ and $N_7$ are more sensitive (iii) \textit{Case 4:} $N_5$ \&$N_6$ are less sensitive and $N_7$ is more sensitive (iv) \textit{Case 5:} $N_5$ \&$N_6$ are more sensitive and $N_7$ is less sensitive (v) \textit{Case 6:} $N_5$ \&$N_7$ are less sensitive and $N_6$ is more sensitive.
 Based on the aforementioned experiments we conclude that making $N_5$ and $N_6$ more sensitive increased the tracking efficiency of the worm but worsened the foraging performance. Changing the sensitivity of $N_7$ did not affect the performance as much. Based on the results of our simulations we conclude that even a $10\%$ drift in the network connectivity affect the  performance only marginally - the worst case foraging efficiency is $72.5\%$, and the worst case deviation during tracking is about $5\%$. This is a significant advantage of our network architecture, illustrating its noise-resilience characteristics.

In order to identify the merits of our spiking-neuron based network configuration, we evaluated the performance of our network against a non-spiking network model with network structure and navigational strategies similar to our spiking-network configuration (Detailed description of model is provided in SI). In the 200 simulations performed, the set-point was identified in $69\%$ of the cases within $1500\,$s as opposed to  $92\%$ in the case of the spiking-neuron model. Further, the average time needed to identify the set-point was $561.67\,$s with a standard deviation of $368.41\,$s. The deviation while tracking the set-point has an average value of $6.2338\,$mM, which is $\approx 10\%$ of the total range of concentrations in the arena, and a standard deviation of $1.7419\,$mM, which is $2.9\%$ of the total range. Compared to the performance of the spiking network which was able to track the set-point with mean deviation of about $1\%$, the non-spiking network's performance deteriorates significantly. When the worm guided by the non-spiking network configuration is exposed to a noisy environment, while tracking the contour, the deviation from the set-point has an average value of $7.5930\,$mM, which is $\approx 12.7\%$ of the total range of concentrations in the arena and a standard deviation of $2.5918\,$mM ($4.32\%$). Compared the spiking network, which shows a mean deviation of $2.8\%$ while tracking the set-point even in the prescence of noise, it is clear that the non-spiking network configuration is less noise-resilent. As is evident from the above results, the spiking-neuron network is much more powerful in terms of both efficiency in detecting the target set-point and tracking it in noisy environments. 

\begin{figure}[!t]
\centering
\includegraphics[width=4in]{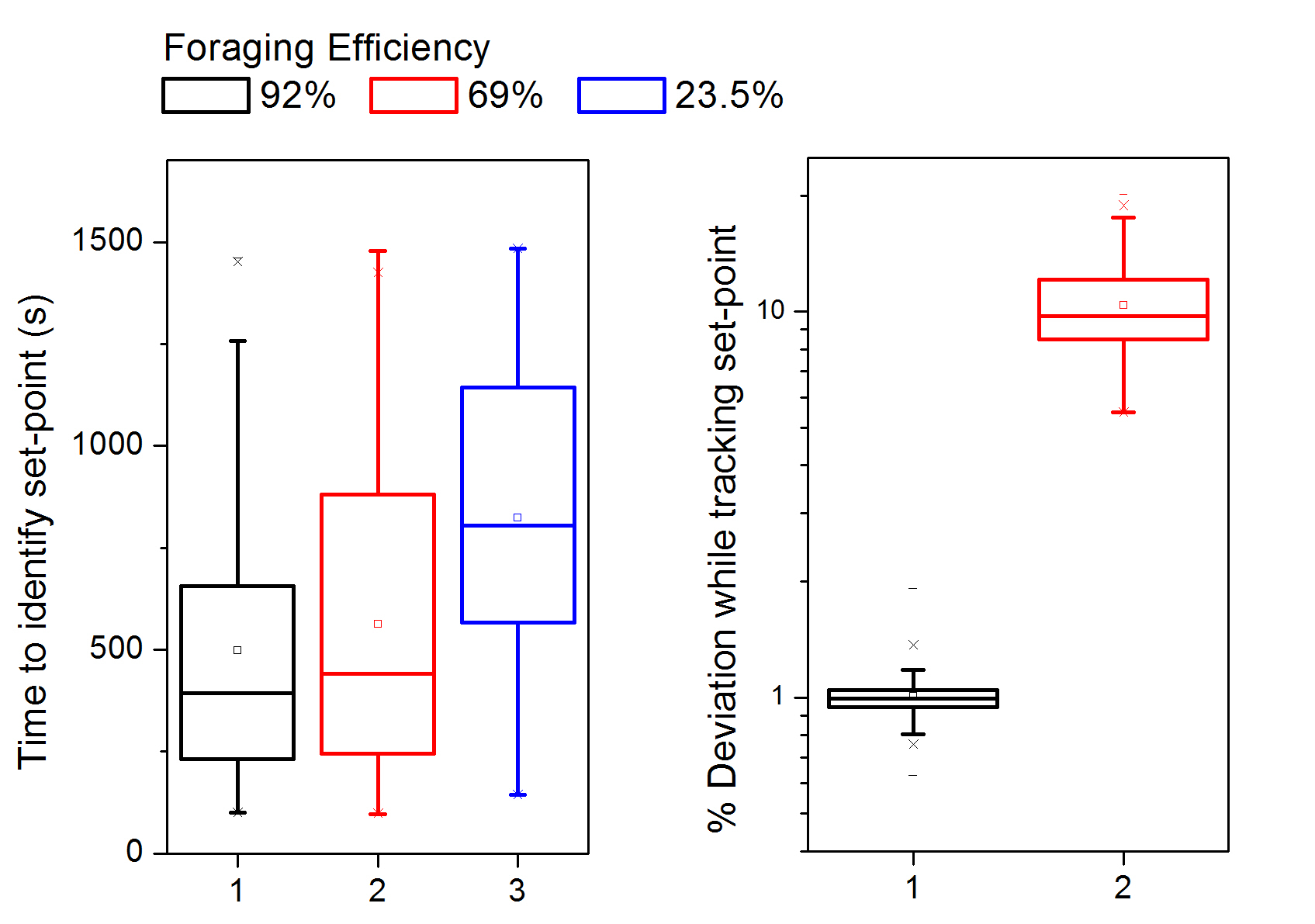}
\caption[Evaluation of network performance against analogous non-spiking network and optimal memory-less  L\'{e}vy foraging strategy ]{ Performance evaluation of foraging and tracking behaviour of our network, (baseline in black) in comparison to - (i) \textit{Red:} tracking and foraging behaviour of a non-spiking neuron network for contour tracking and navigation (ii) \textit{Blue:}  foraging behaviour resulting from L\'{e}vy distribution.
\textit{Left:} Time to identify set-point for the different network configurations \textit{Right:} \% Deviation (with respect to range of concentrations in the arena) while tracking set-point for different network configurations. Also shown in the figure is the foraging efficiency (percentage of 200 simulations where set-point is successfully identified) for the different network configurations.}
\label{pe_basic}
\end{figure}

We compared our foraging strategy to the optimal search strategy for finding randomly distributed targets, where flight-lengths between random turns follow the heavy-tailed L\'{e}vy distribution \cite{nolan:2015}, in order to evaluate the performance of our foraging strategy. In our simulations, run lengths were selected from a truncated L\'{e}vy distribution with $P(l) \propto l^{-2}$ with $l$ belonging to the interval $[s_{min}, s_{max}]$. The values for $s_{min}$ and $s_{max}$ were determined empirically from the neuron model. $s_{min}=0.2649\,$mm was chosen as the most probable run-length and $s_{max} = 40\,$mm as the maximum flight-length for the neuron model. The set-point was reached in only $23.5\%$ of the cases within $1500\,$s as opposed to  $92\%$ in the case of the neuron model.  For these cases, the average time needed to identify the set-point is $824.10\,$s and the standard deviation is $380.72\,$s. For both the models - our neuron model and L\'{e}vy model, the success criteria for foraging was set as the worm reaching within $0.5\,$mM of the set-point. Figure \ref{pe_basic} compares the performance of our spiking-neuron network model to the aforementioned non-spiking network model and the L\'{e}vy foraging strategy. 

Based on the results of the aforementioned experiments, it is evident that our model is able to perform extremely reliable and noise-resilient navigation requiring minimal computational modules and only one single sensor input. Our results thus quantitatively provide a model where sophisticated functions can be performed based on rather simple circuits, as is the case in  \textit{C.elegans}. Our network is a power-efficient and simple alternative for many basic components of robotic navigation. Even though we deploy a simple algorithm for exploration, its power is highlighted when compared against the memoryless L\'{e}vy foraging, which is considered to be the most optimal strategy to detect sparsely distributed targets in unknown environments using local information alone.  Our model, which implements a very rudimentary form of short-term memory, achieved through the adaptation mechanisms that we have mimicked in the ASEL and ASER neurons, presents a foraging strategy which is $\approx 4$ times more efficient than the L\'{e}vy foraging strategy. This enhancement in performance is due to the fact that our model permits long distance explorations while foraging in low-gradient environments, and at the same time, steers it away from incorrect directions when such gradients are encountered. 

The simulations showing the behaviour of our network  in a noisy environment and in the presence of variations in the synaptic weight demonstrate that our network is very resilient to noise at the sensor level as well as at the architecture level. To the best of our knowledge, such reliable form of navigational ability based on  minimal sensory information and computation has not been demonstrated by any other similar algorithms. Further, our comparison of the spiking and non-spiking models demonstrate the importance of the spiking neurons in the network.  In the non-spiking model, the analog potential of the neurons would be similar to the average activity of the spiking neurons. Despite this, the performance of the non-spiking network is significantly worse than the spiking model demonstrating the importance of precisely timed spikes in ensuring robust behaviour.

\begin{figure}[t!]
        \centering
        \begin{subfigure}[b]{0.5\textwidth}
                \includegraphics[width=3in]{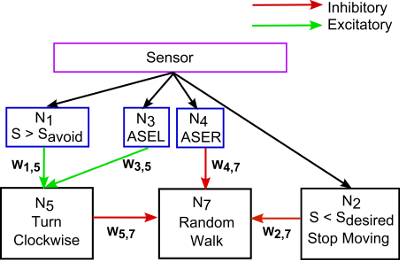}
                \caption{}
                \label{neurons_oa}
        \end{subfigure}%
\hfill
        ~ 
        \begin{subfigure}[b]{0.45\textwidth}
                \includegraphics[width=2.5in]{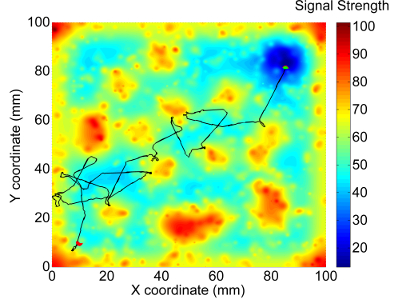}
                \caption{}
                \label{obst}
        \end{subfigure}
        \caption[Modified network for target location and obstacle avoidance]{(a) Schematic diagram of obstacle avoidance network. $N_1$ spikes if worm is too close to obstacles, i.e., S is more than $S_{avoid}$. $N_3$ and $N_4$ are positive and negative gradient detectors respectively.$N_5$ controls deterministic turns in the clockwise direction when the worm is too close to obstacles and is moving even closer while $N_7$ controls random walk. $N_2$ spikes if proximity to destination is less than minimum desired or $S < S_{desired}$ in which case the worm stops moving as it has reached it\rq{}s destination. (b) Track for modified network for obstacle avoidance in an extremely noisy environment. All the hills indicate obstacles and the valley represent destination. The worm starts out in the bottom left corner with initial position indicated by the red dot. It is asked to locate its destination represented by signal strength of $20$ units or less while avoiding obstacles represented by signal strength of $65$ units or more. As can be seen from the figure, the worms finds the destination, in the top right corner, while successfully avoiding all the obstacles. This navigation and obstacle avoidance network functions as desired despite the presence of large amounts of noise. Final position of worm is indicated by green dot.}\label{fig:expnum}
\end{figure}

\subsection{Navigation and obstacle avoidance}

The designed neural network could also be used as an obstacle detector with a simple modification. We consider an arena with the hills representing obstacles and the valleys representing desired locations. The worm uses two parameters - $S_{avoid}$ and $S_{desired}$ to guide it\rq{}s movements. When signal sensed, $S$ is greater than some $S_{avoid}$ it implies that the worm is closer to the obstacle than desired and it must turn. $S_{desired}$ represents the minimum proximity the worm must have to the destination. When the worm is sensing signal $S \leq S_{desired}$, it has reached it\rq{}s destination and can stop moving.

The primary considerations while modifying the contour tracking network to perform obstacle avoidance are: \\
\textbf{1.} The $N_1$- $N_3$- $N_5$ sub-network is exactly identical to the contour tracking case. $N_1$ spiking indicates $S > S_{avoid}$ and $N_3$ spiking indicates the worm is moving closer to the obstacle. When both $N_1$ and $N_3$ spike, $N_5$ spikes and the worm makes a deterministic clockwise turn. The velocity is set to $v_2 = 0.04\,$mm/s by the spiking of $N_5$ to ensure good performance as done in the contour tracking network. \\
\textbf{2.} When the worm is moving down a gradient, indicated by $N_4$ spiking, it signifies it is moving closer to the destination. In this case the course of the worm should be unaltered. Hence random walk must not occur in this case.\\
\textbf{3.} When the worm is sensing a signal $S\leq S_{desired}$, it must stop moving. Therefore as soon as $N_2$ spikes, the worm motion is terminated.\\
\textbf{4}. The worm must randomly explore territory as long as the scenarios portrayed in 1, 2 \& 3 do not occur, i.e, $N_2$, $N_4$ and $N_5$ are not spiking. For obstacle avoidance the random walk neuron $N_7$ is biased with a positive bias current $I_{bias,7}  = 1.36\,$nA and is connected to $N_2$, $N_4$ and $N_5$ with inhibitory synapses with $w = -1$ for all three synapses. The velocity for random walk is set to $v_1 = 0.3\,$mm/s by spiking of $N_7$ to permit random exploration just as in the contour tracking case. The turn angle for random walk  is reduced in this case to regulate performance. When $N_7$ spikes, the worm will make a random turn with an angle in the range of [$-15^\circ$, $15^\circ$].

 The obstacle avoidance network uses $6$ out of the $7$ neurons required for contour tracking. The worm turns only based on spiking of $N_5$ or $N_7$ and stops when $N_2$ spikes. In all other cases it continues along the same path. Figure \ref{neurons_oa} shows the modified network for performing obstacle avoidance. Figure \ref{obst} shows a sample track of the worm while avoiding obstacles. The worm starts out on a roughly flat surface with S of about $40$ units and it is supposed to avoid $S > S_{avoid} = 65$ units  and reach $S \leq 20$ units. 


\section{Discussion}

\ In this paper, we developed a spiking neural network (SNN) for contour tracking and navigation inspired by the chemotaxis network of the nematode \textit{C. elegans}. SNNs have great potential due to their ability to offer, at the very least, the same computational power as their non-spiking counterparts, and for some tasks need lesser number of neurons. 
Our objective was to develop a bio-inspired circuit that could find applicability in real-world robotic applications while harnessing all the  advantages of SNNs. Our network, which contains merely $7$ spiking neurons, relies on input from a sole concentration sensor and exhibits the ability to forage and track a desired concentration set-point using only binary values or spikes for all information transmission. 

We first modeled a pair of chemosensory neurons to act as positive and negative gradient detectors, motivated by the ASE neuron pair in \textit{C. elegans}. The output of these neurons was translated into spikes by simple thresholding, which was then processed by downstream neurons to make decisions using navigational strategies inspired by the \textit{klinokinesis} of  \textit{C. elegans}.
Our simulations show that the spiking neural network configuration, when compared with an equivalent non-spiking neural network that used graded potentials  has $\approx 1.33$ times higher chances of identifying the set-point and tracks the set-point with $10$ times more efficiency. With the same number of neurons and navigational strategies in both networks, we show that our network is not only more efficient, but also more noise tolerant.

Our simulations show that our worm is able to detect the set-point with $\approx 4$ times higher probability than the optimal memoryless L\'{e}vy foraging model, with the average foraging time $\approx 0.6$ times that needed by the L\'{e}vy foraging strategy. Once the worm reaches the desired set-point, it tracks it with an average standard deviation of $\approx$ 1\% of the range of concentration in the arena. In earlier work by \cite{Pappleby2013} (which utilized graded potential model for the ASE neurons), introduction of even modest levels of noise completely disrupts the chemotactic drive of the simulated worm. Our model which is based on spiking neurons  shows great resilience to noise and even with high degrees of salt and pepper noise, with peak magnitude as high as $20\%$ of the range of concentrations in the arena, our \lq\lq{}worm\rq\rq{} is able to track the set-point with an average deviation of about $3\%$ the range of concentrations in the arena. This offers great engineering potential as all real world applications requiring motor control based on sensory information would be prone to environmental noise and sensory errors.  We also show that our network is able to perform well in the presence of drift in the network configuration, especially the synaptic weights. Our experiments show that even when the weights shift by as much as $10\%$, the foraging efficiency of the worm  is about $72.5\%$ (compared to the optimal efficiency of $92\%$). This further strengthens the practical applicability of our network.

The developed network could be used for navigation control guided by concentration or intensity of any environmental variable such as temperature, radiation, noxious gases etc by choosing appropriate sensors and is by no means confined to chemical concentration. Our network could be adapted to operate in various temporal and spatial scales, with very minor modifications, which is pertinent while designing systems for robotic motion, autonomous flight control, etc, making the network very versatile. Since our network uses only temporal gradients to determine future exploration decisions, it is possible to extend its performance to three dimensions in a straight-forward manner. The network performance in oblivious to whether to new velocity vector chosen is constrained to two dimensions or whether it can extend to all three dimensions.  The practical applicability of our worm is not confined to contour-tracking and we have shown that with slight modifications, our network could also be used to guide robotic motion while avoiding obstacles. Since we have identified scaling rules to translate its operation to other spatial and temporal domains, while keeping the average spike rate of the neurons close to the $1-10$ Hz seen in biological systems, extremely energy efficient hardware implementations are also possible. We believe that the developed algorithm could offer a potentially more robust, noise-resilient and energy efficient alternative to conventional motor control algorithms in a wide variety of engineering applications.

\bibliographystyle{abbrv} 
\bibliography{test}

\end{document}